\documentclass[letterpaper]{article}
\usepackage{aaai19}
\usepackage{amsfonts,amsmath}
\usepackage{graphicx}
\usepackage{times}
\usepackage{helvet}
\usepackage{courier}
\usepackage{booktabs,multirow,array,color}
\begin{document}
%
\title{Multi-Task Learning with Multi-View Attention for Answer Selection \\ and Knowledge Base Question Answering}
\author{Yang Deng$^{1}$, Yuexiang Xie$^{1}$, Yaliang Li$^2$, Min Yang$^3$, Nan Du$^2$, Wei Fan$^2$, Kai Lei$^{1*}$, Ying Shen$^{1*}$\\
    $^*$Corresponding Author \\
  $^1$Shenzhen Key Lab for Information Centric Networking \& Blockchain Technology (ICNLAB),\\
  School of Electronics and Computer Engineering, Peking University Shenzhen Graduate School \\
  $^2$Tencent Medical AI Lab $^3$Shenzhen Institutes of Advanced Technology, Chinese Academy of Sciences\\ 
  {\tt \{ydeng,xieyx\}@pku.edu.cn, yaliangli@tencent.com, min.yang@siat.ac.cn,}\\
  {\tt \{ndu,davidwfan\}@tencent.com, \{leik,shenying\}@pkusz.edu.cn}}

\maketitle

\begin{abstract}
Answer selection and knowledge base question answering (KBQA) are two important tasks of question answering (QA) systems. Existing methods solve these two tasks separately, which requires large number of repetitive work and neglects the rich correlation information between tasks. 
In this paper, we tackle answer selection and KBQA tasks simultaneously via multi-task learning (MTL), motivated by the following motivations. First, both answer selection and KBQA can be regarded as a ranking problem, with one at text-level while the other at knowledge-level. Second, these two tasks can benefit each other: answer selection can incorporate the external knowledge from knowledge base (KB), while KBQA can be improved by learning contextual information from answer selection. 
To fulfill the goal of jointly learning these two tasks, we propose a novel multi-task learning scheme that utilizes multi-view attention learned from various perspectives to enable these tasks to interact with each other as well as learn more comprehensive sentence representations. 
The experiments conducted on several real-world datasets demonstrate the effectiveness of the proposed method, and the performance of answer selection and KBQA is improved. Also, the multi-view attention scheme is proved to be effective in assembling attentive information from different representational perspectives. 
\end{abstract}

\section{Introduction}
Question answering (QA) is an important but challenging NLP application. Nowadays, there are many possible sources of data for the QA systems, such as web documents, QA communities, knowledge bases and so on. According to these data sources, question answering can be divided into several different tasks, including machine reading comprehension~\cite{DBLP:conf/emnlp/RajpurkarZLL16}, answer selection~\cite{Wang2007What}, knowledge base question answering~\cite{Bordes2015Large} and more. Recent years have witnessed many successes in applying deep neural networks on these QA tasks~\cite{dos2016attentive,Yin2016simple,Hao2017An}.

Despite the advancement of these models, different QA tasks are still solved separately. Designing and training various models for specific tasks are time-consuming and expensive. Recently, as seen in many other NLP tasks, multi-task learning has been extensively studied to learn multiple related tasks simultaneously. The applications of MTL are extensive in NLP domain, such as text classification~\cite{DBLP:conf/ijcai/ZhengCQ18}, sequence labeling~\cite{DBLP:conf/aaai/ChenQLH18} and text summarization~\cite{guo2018soft}, etc. However, applying MTL on QA has received little attention.

In this work, we explore multi-task learning approaches to tackle answer selection and knowledge base question answering at the same time, with the intuition that these tasks both can be regarded as a ranking problem, but one in text-level and the other in knowledge-level. Specifically, the task of answer selection aims to pick out the correct answers for the given question from a set of candidate answer sentences, while the task of KBQA focuses on extracting the corresponding facts from KB, such as Freebase~\cite{bollacker2008freebase}, to answer the given question. Besides, previous works prove that answer selection task can be benefited from external knowledge~\cite{Savenkov2017EviNets,DBLP:conf/coling/DengSYLDFL18,Shen2018Knowledge}, while incorporating text information also enhances the performance of KBQA task~\cite{yu2017improved,DBLP:conf/coling/SorokinG18}. 

\begin{table*}
\centering
\fontsize{9}{10}\selectfont
\begin{tabular}{cccc}
\toprule
&&Answer Selection&KBQA\\
\midrule
\multirow{2}{*}{word}&Q&what was johnny appleseed 's real name ? &what is the name of a track created by katy perry ?\\
&A&john chapman , aka american folk hero johnny appleseed .&katy perry music artist track witness\\
\multirow{2}{*}{knowledge}&Q& johnny\_appleseed& katy\_perry \\
&A& john\_chapman, johnny\_appleseed&katy\_perry, music.artist.track, witness \\
\bottomrule
\end{tabular}
\caption{\label{table1} Examples of Answer Selection and KBQA Data}
\end{table*}

Most existing multi-task learning schemes divide the layers of a model into task-specific and shared layers~\cite{DBLP:conf/icml/KumarD12,guo2018soft}. The shared layers are shared across all tasks, while the task-specific layers are separate for each task. However, these methods neglect the interrelation between the task-specific layers and the shared layers, and the interaction among different tasks. Thus, we present a novel multi-task learning scheme to learn multi-view attentions from different aspects, which enables different tasks to interact with each other. Concretely, we assemble the attentive information from the task-specific layers to learn more comprehensive sentence representations in the shared layers. In addition, the multi-view attention mechanism enhances the sentence representational learning by combining word-level and knowledge-level information. That is to say, the attentive information in both word-level and knowledge-level are shared and transferred among different tasks by using the multi-view attention scheme.

To demonstrate the effectiveness of the proposed method, we conduct experiments on both answer selection and KBQA datasets. Empirically, joint learning of answer selection and KBQA tasks significantly improves the performance of each task compared to learning them independently. The experimental results also indicate the effectiveness of the multi-view attention scheme and each view‘s attention contributes.

In summary, our main contributions are as follows: 
\begin{itemize}
\item We explore multi-task learning approaches for answer selection and knowledge base question answering. Answer selection task can be improved by KBQA task in knowledge-level, while KBQA task can be enhanced by answer selection task in word-level.
\item We propose a novel multi-task learning scheme that leverages multi-view attention mechanism to bridge different tasks, which integrates the important information of the task-specific layers into the shared layers as well as enables the model to interactively learn word-level and knowledge-level representations.
\item Experimental results show that multi-task learning of answer selection and KBQA outperforms state-of-the-art single-task learning methods. Besides, the multi-view attention based MTL scheme further enhance the performance.
\end{itemize}

\section{Multi-Task Learning for Question Answering}
In this section, we introduce the multi-task learning of answer selection and knowledge base question answering.

\subsection{Problem Definition}
The tasks of answer selection and KBQA can be typically regarded as a ranking problem. Given a question $q_i\in Q$, the task is to rank a set of candidate answer sentences or facts $a_i\in A$. Specifically, a function $f(q, a)$ that computes a relevancy score $f(q, a)\in [0, 1]$ for each question-answer pair. 

The multi-task learning of answer selection and KBQA starts with the entity linking results. As the example shown in Table ~\ref{table1}, a word sequence $W=\{w_1,w_2,...,w_L\}$ and a knowledge sequence $K=\{k_1,k_2,...,k_L\}$ are prepared for each question and each candidate answer. For the question and answer in answer selection and the question in KBQA, we derive the knowledge of the sentence by entity linking~\cite{Savenkov2017EviNets}. For the answer fact in KBQA, we obtain the word sequence from the tokenized entity name and relation name~\cite{yu2017improved}.

In the multi-task scenario, we aim to rank the candidate answers for each question from $T$ related tasks. We refer $D_t$ as the $t$-th preprocessed task dataset with $N$ samples:
\begin{equation}
 D_t = \{(W^{(t)}_{q_i},K^{(t)}_{q_i},W^{(t)}_{a_i},K^{(t)}_{a_i},Y^{(t)}_i)\}^{N_t}_{i=1},
\end{equation}
where $Y^{(t)}_i$ denotes the label of $i$-th QA pair in $t$-th task.

\subsection{Multi-Task Question Answering Network}
The basic multi-task learning model is a deep neural network that adopts layer-specific sharing mechanism~\cite{guo2018soft} which shares some high-level information across different tasks and the remained layers are parallel and independent to learn task-specific low-level information. 
\begin{figure}
\centering
\includegraphics[width=0.38\textwidth]{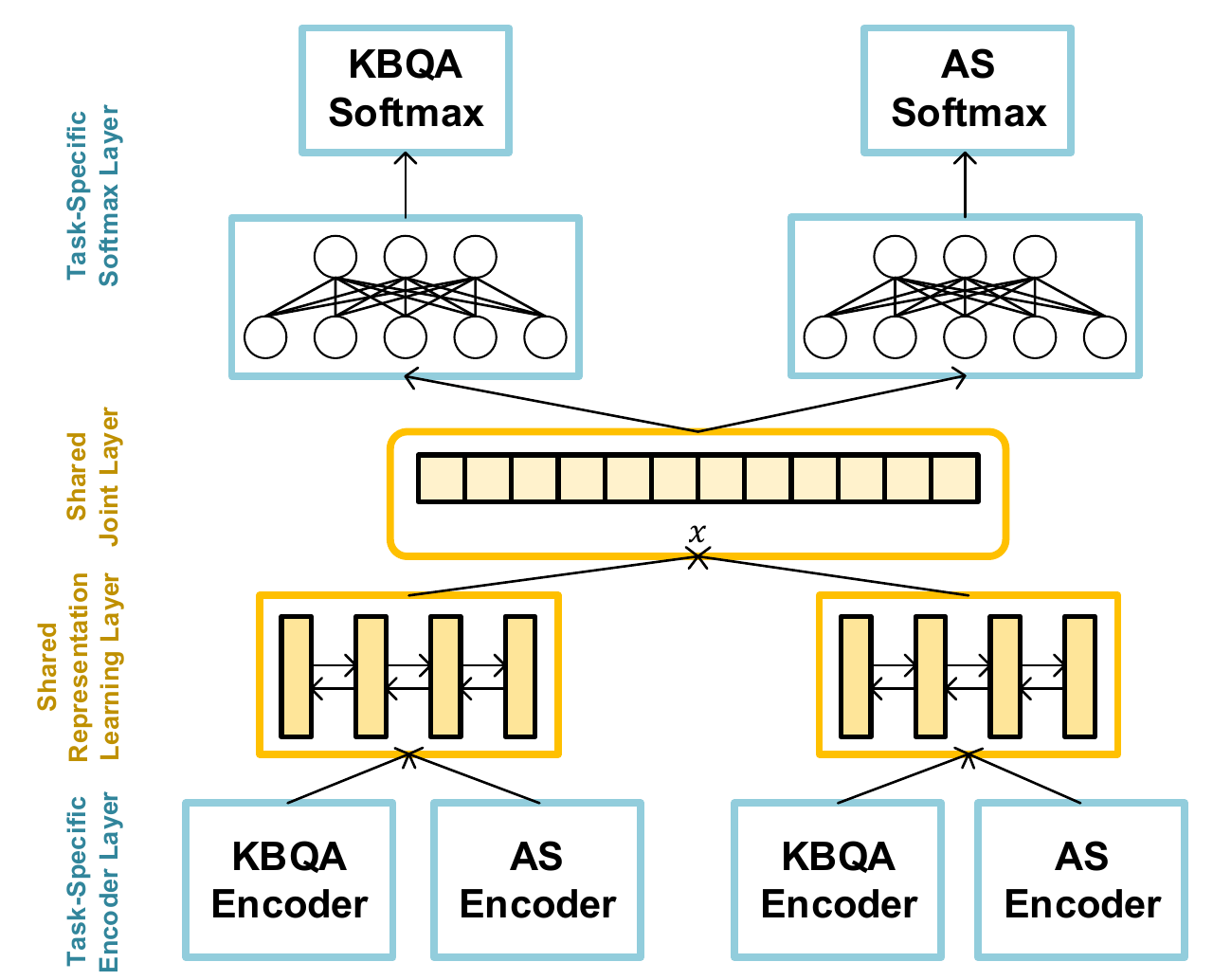}
\caption{Basic Multi-Task QA Network}
\label{figure1}
\end{figure}
Figure~\ref{figure1} illustrates the overall architecture of multi-task QA network (MTQA-net) for answer selection (AS) and knowledge base question answering (KBQA).

\subsubsection{Task-specific Encoder Layer} 
\begin{figure}
\centering
\includegraphics[width=0.26\textwidth]{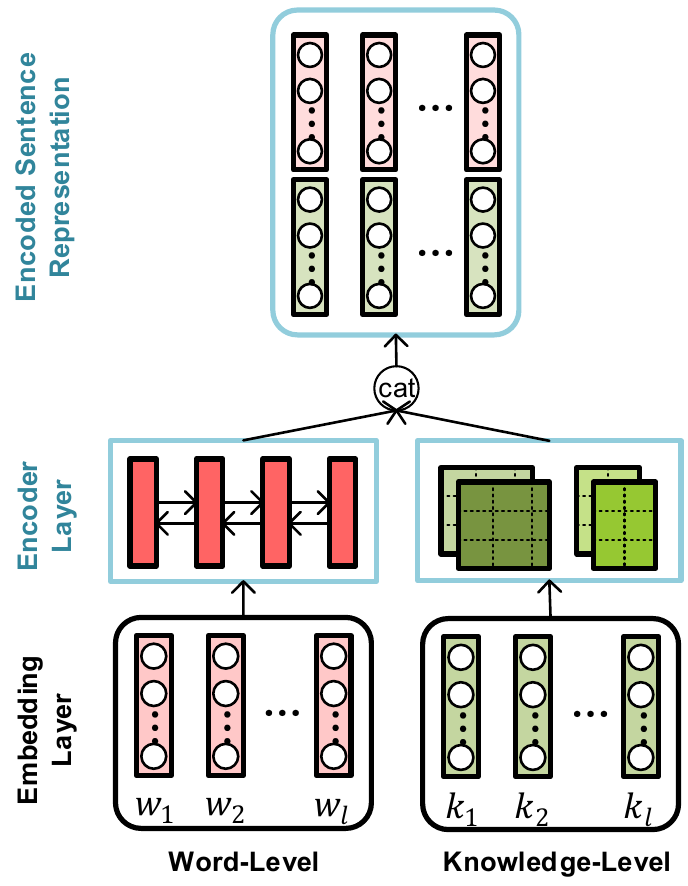}
\caption{Task-specific Encoder Layer}
\label{figure2}
\end{figure}

The preprocessed sentences are first encoded into distributed vector representations. Different QA tasks are supposed to be diverse in data distributions and low-level representations. Therefore, each task is equipped with a task-specific siamese encoder for both questions and answers, and each task-specific encoder contains a word encoder and a knowledge encoder to learn the integral sentence representations, as shown in Figure~\ref{figure2}.

\textbf{Word Encoder.}
The input of the word encoder module is a sequence of word embeddings $E_W=\{e_{w_1},e_{w_2},...,e_{w_L}\}$. We employ Bi-LSTM (Bidirectional Long Short-Term Memory Networks) to capture the context information from both the head-to-tail and the tail-to-head context. The output at $l$-th word is represented by $h_l=[\overrightarrow{h_l}:\overleftarrow{h_l}]$, in which $\overrightarrow{h_l}$ is the output of the forward network and $\overleftarrow{h_l}$ is that of the backward network.
Given a pair of word sequences of the question $q$ and the answer $a$, we generate the word-based sentence representation $H_{W}\in\mathbb{R}^{L\times{d_h}}$ for both the question and the answer, where $L$ and $d_h$ are the length of sentences and the size of hidden units:
\begin{equation}
H_{W_q} = \textbf{Bi-LSTM}(E_{W_q});  H_{W_a} = \textbf{Bi-LSTM}(E_{W_a}).
\end{equation}

\textbf{Knowledge Encoder.} 
Different from word encoder, a sequence of knowledge embeddings $E_K=\{e_{k_1},e_{k_2},...,e_{k_L}\}$ is input into knowledge encoder. As the knowledge sequence is composed by a series of tokenized entity or relation name, the high-level knowledge-based representations are desired for the latter learning procedure. We deal with this problem by applying CNN on the knowledge sequences, in which filters of size $n$ slide over the knowledge embedding matrix to capture the local n-gram features. Each move computes a hidden layer vector as
\begin{align}
 x_l = & [ e_{k_{l-\frac{n-1}{2}}},\ldots,e_{k_l},\ldots, e_{k_{l+\frac{n-1}{2}}}],\\
& h_l = \tanh\left(W_{c}x_l+b_{c}\right),
\end{align}
where $W_c$ and $b_c$ are the convolution kernel and the bias vector to be learned. 

Due to the uncertainty of the length of entities, a couple of filters of various sizes are employed to obtain different output vectors $\left\{H^{\left(1\right)},H^{\left(2\right)},\ldots,H^{\left(n\right)}\right\}$, where $H^{\left(i\right)}$ denotes the output vector obtained by the $i$-th filter. We pass these output vectors through a fully-connected layer to get the knowledge-based sentence representation $H_{K}\in\mathbb{R}^{L\times{d_f}}$, where $L$ is the length of the sentence and $d_f$ is the total filter sizes of CNN. Given the question $q$ and the answer $a$, the knowledge-based sentence representations are:
\begin{align}
H_{K_q} = [H^{\left(1\right)}_{K_q}:H^{\left(2\right)}_{K_q}:\ldots:H^{\left(n\right)}_{K_q}], \\
H_{K_a} = [H^{\left(1\right)}_{K_a}:H^{\left(2\right)}_{K_a}:\ldots:H^{\left(n\right)}_{K_a}].
\end{align}

After obtaining the word-based and the knowledge-based sentence representations, $H_{K_q}$ and $H_{K_a}$ are still vectors in the order of words in the sentence, since the fully-connected layer concatenates the outputs from all the filters in the dimension of the feature instead of the sequence. Thus, the order of $H_{K_q}$ and $H_{K_a}$ is consistent with $H_{W_q}$ and $H_{W_a}$. Then we concatenate them into the encoded sentence representations, $H_{q} = [H_{W_q}:H_{K_q}]$ and $H_{a} = [H_{W_a}:H_{K_a}]$.

\subsubsection{Shared Representation Learning Layer}
After encoding sentence into vector representations with task-specific encoder, we share high-level information across different tasks via a shared representation learning layer. Compared with the input of task-specific encoder layer, the integral sentence representation contains richer semantic meaning and share more similar distributions with other tasks. Therefore, we integrate the encoded vectors from all the tasks and pass through a high-level shared Siamese Bi-LSTM to generate the final QA representations:
\begin{equation}
S_{q} = \textbf{Bi-LSTM}(H_q); \quad S_{a} = \textbf{Bi-LSTM}(H_a).
\end{equation}

We apply average pooling over the Bi-LSTM output, $s_q = Average(S_q)$, $s_a = Average(S_a)$. Inspired by previous works~\cite{Severyn2015Learning} and \cite{Tay2017Learning}, we incorporate some word and knowledge overlap features $x_{ol}\in\mathbb{R}^6$ to form the final feature space for binary classification, including word overlap score, non-stop word overlap score, weighted word overlap score, non-stop weighted word overlap score, knowledge overlap score and weighted knowledge overlap score. Thus, the final feature space will be $x=\left[s_q,s_a,x_{ol}\right]$.

\subsubsection{Task-specific Softmax Layer}
For a question-answer pair, $q_i^{(t)}$ and $a_i^{(t)}$, and its label $y_i^{(t)}$ in $k$-th task, the final feature representations is fed into the task-specific softmax layer for binary classification: 
\begin{equation}
p^{(t)}=\text{softmax}\left(W^{(t)}_sx+b^{(t)}_s\right),
\end{equation}
where $p^{(t)}$ is the predicted probability, $W^{(t)}_s\in\mathbb{R}^{d_x\times{2}}$ and $b^{(t)}_s\in\mathbb{R}^2$ are the task-specific weight matrix and bias vector in the hidden layer. 

\subsection{Multi-Task Learning}
The overall multi-task learning model is trained to minimize the cross-entropy loss function:
\begin{small}
\begin{equation}
 L=-\sum_{t=1}^T \lambda_t \sum_{i=1}^{N_t}\left[y^{(t)}_i\log{p^{(t)}_i}+\left(1-y^{(t)}_i\right)\log{\left(1-p^{(t)}_i\right)}\right],
\end{equation}
\end{small}where $\lambda_t$ is a parameter that determines the weight of $t$-th task, and $y^{(t)}_i$ is the ground-truth label of question-answer pair $(q_i^{(t)}, a_i^{(t)})$. In practice, the same weight is assigned to all tasks.

\section{Multi-Task Model with Multi-View Attention}
In order to enhance the interaction between different QA tasks in latent representation space, we propose a multi-view attention mechanism to fetch the important information from task-specific layers as well as the shared layers. 
\begin{figure*}
\centering
\includegraphics[width=0.6\textwidth]{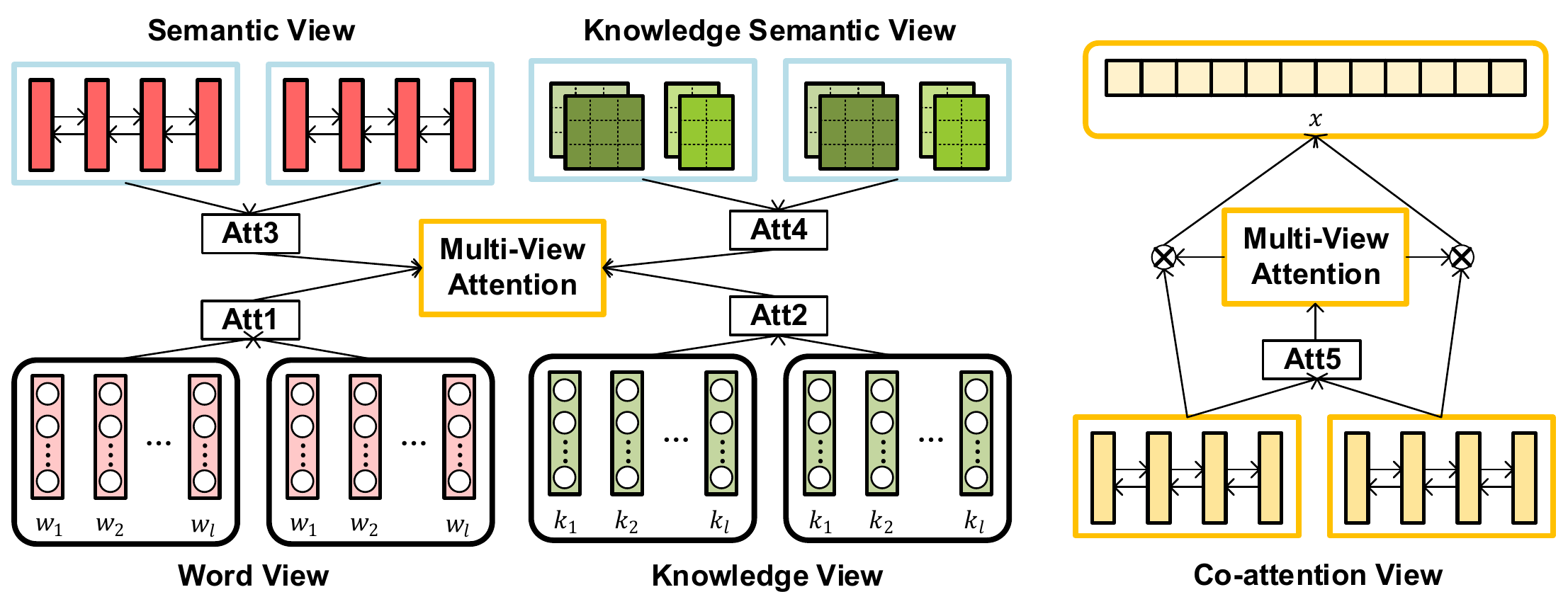}
\caption{Multi-View Attention}
\label{figure3}
\end{figure*}
\subsection{Multi-View Attention Scheme}
As shown in the Figure~\ref{figure3}, different with other attention sharing scheme, we not only draw attention from the task-specific layers, but also combine the information from the shared layers. In addition, we obtain the attentive information from both word and knowledge perspectives, since word-level and knowledge-level information may make a joint contribution to representational learning. Specifically, we compute five views of attention, including the views of word, knowledge, semantic, knowledge semantic and co-attention.
\subsubsection{Word View \& Knowledge View} We first adopt a two-way attention mechanism~\cite{dos2016attentive} to interactively modeling a pair of QA sentences, by attending correspondingly important information in two sentences. We collect the attention weights from both the word and the knowledge embeddings:
\begin{small}
\begin{equation}
 M_{W}=\text{tanh}\left(E_{W_q}^TU_{W}E_{W_a}\right);
 M_{K}=\text{tanh}\left(E_{K_q}^TU_{K}E_{K_a}\right),
\end{equation}
\end{small}where $U_{W}\in\mathbb{R}^{d_{e_w}\times{d_{e_w}}}$ and $U_{K}\in\mathbb{R}^{d_{e_k}\times{d_{e_k}}}$ are parameter matrices to be learned; $d_{e_w}$ and $d_{e_k}$ are the dimensions of word embeddings and knowledge embeddings. 

Then we apply max-pooling over $M_W$ and $M_K$ row-wise and column-wise to obtain the word view and the knowledge view attention weights for the question and the answer:
\begin{small}
\begin{align}
 \alpha_q^{(1)} =\text{softmax}(\text{Max}(M_W)); \alpha_a^{(1)} =\text{softmax}(\text{Max}({M_W}^T)),\\
 \alpha_q^{(2)} =\text{softmax}(\text{Max}(M_K)); \alpha_a^{(2)} =\text{softmax}(\text{Max}({M_K}^T)),
\end{align}
\end{small}where $\alpha_q^{(1)}$ and $\alpha_a^{(1)}$ are the attention weights for the question and the answer from word view; $\alpha_q^{(2)}$ and $\alpha_a^{(2)}$ are the attention weights from knowledge view.

\subsubsection{Semantic View \& Knowledge Semantic View}
As the context semantic information is of great importance in sentence representational learning, we exploit the overall context semantic information as the attention source over all the elements in the sentence. Thus, we apply max/mean pooling over the output of the task-specific encoder layer to obtain the overall semantic information of a sentence. We experimented on using either max or mean pooling to generate the semantic information. The result shows that the following pooling methods achieve the best performance:
\begin{equation}
 o_{w_q} =\text{Average}\left(H_{W_q}\right); \quad o_{w_a} =\text{Average}\left(H_{W_a}\right),
\end{equation}
\begin{equation}
  o_{k_q} =\text{Max}\left(H_{K_q}\right); \quad o_{k_a} =\text{Max}\left(H_{K_a}\right),
\end{equation}

Conceptually, the attention mechanism takes into consideration the semantic information,  which is expected to capture the correlations between question words and answer words:
\begin{small}
\begin{align}
 \alpha_q^{(3)}=\text{softmax}(w_{w_q}^T\text{tanh}(W_{w_a}o_{w_a}+W_{w_q}H_{W_q})),\\
 \alpha_a^{(3)}=\text{softmax}(w_{w_a}^T\text{tanh}(W_{w_q}o_{w_q}+W_{w_a}H_{W_a})),\\
 \alpha_q^{(4)}=\text{softmax}(w_{k_q}^T\text{tanh}(W_{k_a}o_{k_a}+W_{k_q}H_{K_q})),\\
 \alpha_a^{(4)}=\text{softmax}(w_{k_a}^T\text{tanh}(W_{k_q}o_{k_q}+W_{k_a}H_{K_a})),
\end{align}
\end{small}where $W_{w_q}, W_{w_a}\in\mathbb{R}^{d_h\times{d_h}}$, $w_{w_q}, w_{w_a}\in\mathbb{R}^{d_h}$, $W_{k_q}, W_{k_a}\in\mathbb{R}^{d_f\times{d_f}}$, $w_{k_q}, w_{k_a}\in\mathbb{R}^{d_f}$ are attention parameters to be learned; $\alpha_q^{(3)}$ and $\alpha_a^{(3)}$ are the attention weights for the question and the answer from semantic view; $\alpha_q^{(4)}$ and $\alpha_a^{(4)}$ are the attention weights from knowledge semantic view.

\subsubsection{Co-attention View}
Similar to word view and knowledge view attention, we employ a two-way attention mechanism to generate the co-attention between final question and answer representations:
\begin{small}\begin{align}
 M_{co}=\text{tanh}\left(S_q^TU_{S}S_a\right),\\
 \alpha_q^{(5)} = \text{softmax}(\text{Max}(M_{co})),\\
 \alpha_a^{(5)} =\text{softmax}(\text{Max}({M_{co}}^T)),
\end{align}
\end{small}where $U_{W}\in\mathbb{R}^{d_{s}\times{d_{s}}}$ is the attention parameter matrix to be learned; $d_{s}$ is the dimension of final QA representations; $\alpha_q^{(5)}$ and $\alpha_a^{(5)}$ are the co-attention weights for the question and the answer.

\subsubsection{Multi-View Attentive Representation}
We define the multi-view attention fusion from word, knowledge, semantic, knowledge semantic and co-attention views as:
\begin{small}
\begin{equation}
\alpha_q = \text{softmax}\sum_{i=1}^5 \lambda_q^{(i)}\alpha_q^{(i)};
\alpha_a = \text{softmax}\sum_{i=1}^5 \lambda_a^{(i)}\alpha_a^{(i)},
\end{equation}
\end{small}where $\lambda_q^{(i)}$ and $\lambda_a^{(i)}$ are hyper-parameters that determines the weights of the five kinds of attentions. In order to observe the contribution of each view of attention, we assign the same weight to all views of attention in the experiment. Finally, the attentive QA representations will be:
\begin{equation}
s_q = S_q\alpha_q; \quad s_a = S_a\alpha_a.
\end{equation}

\subsection{Multi-View Attention Sharing}
As the multi-view attention is applied over hidden states in the shared representation layer, the parameters to compute the attention weights are supposed to be shared across tasks as well. Meanwhile, different tasks are connected by the multi-view attention, since the multi-view attention scheme gathers the information from both task-specific layers  and the shared layers. 

\section{Experiment}

\subsection{Datasets \& Preprocessing}
We use YahooQA \cite{Tay2017Learning} and TREC QA \cite{Wang2007What} for answer selection task, and SimpleQuestions \cite{Bordes2015Large} and WebQSP \cite{Yih2016The} for knowledge base question answering task. The statistics of these datasets are described in Table~\ref{data}.

\begin{table}[htb]
\fontsize{7}{9}\selectfont
\centering
  \begin{tabular}{cccc}
    \toprule
	Dataset & \#Question (train/dev/test)&\#QA Pair (train/dev/test) \\
    \midrule
    Yahoo QA & 50098/6289/6283 & 253K/31K/31K\\
   TREC QA & 1229/82/100 & 53417/1148/1517\\
   SimpleQuestions & 71038/10252/20464 & 571K/80K/164K \\
   WebQSP & 3067/-/1632 & 302K/-/160K\\
  \bottomrule
\end{tabular}
 \caption{Summary statistics of datasets.\label{data}}
\end{table}

 \emph{YahooQA}\footnote{https://github.com/vanzytay/YahooQA\_Splits} An open-domain community-based dataset collected from Yahoo Answers. \citeauthor{Tay2017Learning} \shortcite{Tay2017Learning} filters out questions and answers whose length is out of 5-50, and generates 4 negative samples for each question. The same metrics as~\cite{tay2017cross,DBLP:conf/coling/DengSYLDFL18} are adopted for evaluation, including Precision@1 and MRR.  

 \emph{TREC QA} A widely-adopted factoid question answering dataset. Following previous works~\cite{tay2017cross,DBLP:conf/coling/DengSYLDFL18}, we experiment on the raw TREC QA dataset and use Mean Average Precision (MAP) and Mean Reciprocal Rank (MRR) as evaluation metrics.

 \emph{SimpleQuestions} A single-relation KBQA dataset. This dataset consists of questions annotated with a corresponding fact from Freebase that provides the answer. We report Accuracy as previous studies~\cite{Yih2016The,yu2017improved}. A question is considered answered correctly only when the predicted positive answers match one of the ground-truths.

 \emph{WebQSP} A multi-relation KBQA dataset. 
Yih et al. (2016) created this dataset by extracting the questions that are answerable using Freebase from WebQuestions~\cite{Berant2014Semantic}. We adopt Accuracy as the evaluation metric as the SimpleQuestions dataset.

In our setting, we assume that the data are preprocessed by entity linking, which means entities in sentences have already been extracted and linked to certain entities in the KB. 

 \emph{Entity Linking}  We use FB5M\footnote{https://research.facebook.com/researchers/1543934539189348} as the knowledge base, which contains 4,904,397 entities, 7,523 relations, and 22,441,880 facts. For YahooQA and TREC QA\footnote{https://github.com/dengyang17/MTQA}, we label all the sentences with the entity mentioned in themselves. We apply the TagMe\footnote{https://github.com/marcocor/tagme-python} entity linker to extract entity mentions from sentences, and only keep the entity mentions with the confidence score above the 0.2 threshold. For each entity mention, we retrieve one certain entity from FB5M. For SimpleQuestions and WebQSP, we start with the entity linking results\footnote{https://github.com/Gorov/KBQA\_RE\_data} as Yu et al. (2017).

\subsection{Experiment Settings}
The word embeddings for all the models are initialized by pre-trained GloVE embeddings\footnote{http://nlp.stanford.edu/data/glove.6B.zip} of 300 dimensions. TransE~\cite{Bordes2013Translating} is adopted as the knowledge embedding method to generate the knowledge embeddings for all the models. OpenKE\footnote{https://github.com/thunlp/OpenKE} is employed to implement TransE with the default settings.

For all the implemented models, we apply the same parameter settings. The LSTM hidden layer size and the final hidden layer size are both set to 200. The
width of the convolution filters is set to be 2 and 3, and the number of
convolutional feature maps is set to be 100. The learning rate and the dropout rate are set to 0.0005 and 0.5 respectively. We train our models in batches with size of 128. All other parameters are randomly initialized from [-0.1, 0.1]. The model parameters are regularized with a L2 regularization strength of 0.0001. The maximum length of sentence is set to be 40.

\subsection{Multi-Task Learning Results}

\begin{table*}[htb]
\fontsize{7}{9}\selectfont
\centering
\begin{tabular}{ccccccc}
\toprule
\multirow{2}{*}{Model}& \multicolumn{2}{c}{Yahoo QA}& \multicolumn{2}{c}{TREC QA}&SimpleQuestions&WebQSP \\ 
\cmidrule(lr){2-3}\cmidrule(lr){4-5}\cmidrule(lr){6-6}\cmidrule(lr){7-7}
&P@1&MRR&MAP&MRR&Accuracy&Accuracy\\
\midrule
HD-LSTM \cite{Tay2017Learning} & 0.557 & 0.735 & 0.750 & 0.815& -& - \\
CTRN \cite{tay2017cross} & 0.601 & 0.755 & 0.771 & 0.838& -& -  \\
HyperQA \cite{Tay2018hyper} & 0.683 & 0.801 & 0.770 & 0.825& -& -  \\
KAN(AP-LSTM) \cite{DBLP:conf/coling/DengSYLDFL18} & \underline{0.744} &\underline{0.840} & \underline{0.797} & \underline{0.850}& -& -  \\
BiCNN \cite{Yih2015Semantic}& -& - & -& -  & 0.900 & 0.777 \\
AMPCNN \cite{Yin2016simple}& -& - & -& -  & 0.913 & - \\
HR-BiLSTM \cite{yu2017improved}& -& - & -& -  & 0.933 & 0.825 \\
Multiple View Matching (Yu et al., 2018)& -& - & -& -  & \underline{0.937} & \underline{0.854} \\
\midrule
MTQA-net (STL) & 0.737 & 0.818& 0.763& 0.832&0.913&0.808\\
MTQA-net (MTL) & 0.752 & 0.839 & 0.779 & 0.841&0.922 &0.820\\
MVA-MTQA-net (STL) & 0.806 & 0.878 & 0.783 & 0.838& 0.931 & 0.823\\
MVA-MTQA-net (MTL) &\textbf{0.833}&	\textbf{0.909} &\textbf{0.811}&\textbf{0.862}& \textbf{0.957} & \textbf{0.858}\\
\bottomrule
\end{tabular}
\caption{\label{table2} Multi-Task Learning Results}
\end{table*}

Table~\ref{table2} summarizes the experimental results of different methods on answer selection and knowledge base question answering. For answer selection task, four methods listed in Table~\ref{table2} achieve the state-of-the-art results in Yahoo QA and TREC QA datasets. The first three methods~\cite{Tay2017Learning,tay2017cross,Tay2018hyper} are traditional single-task learning methods, while \citeauthor{DBLP:conf/coling/DengSYLDFL18} (2018) employs transfer learning method to pre-train the model with a large-scale dataset and leverages external knowledge from KB to improve the sentence representational learning. For KBQA task, we compare the proposed method to four single-task learning state-of-the-art methods. Note that we start with the same entity linking results as \citeauthor{yu2017improved} (2017).

In general, the proposed multi-view attention based MTL method, MVA-MTQA-net (MTL), outperforms the state-of-the-art results by a noticeable margin on all the datasets. For instance, on the YahooQA and SimpleQuestions dataset, the proposed method improves about 8\% and 2\% on the metrics over these baselines. 

In both MVA-MTQA-net and its basic model (MTQA-net), multi-task learning (MTL) methods can significantly improve the performance of all four datasets compared with single-task learning (STL), which demonstrates the effectiveness of combining answer selection and knowledge base question answering to conduct multi-task learning.

\subsection{Ablation Analysis of Multi-View Attention}

\begin{table*}[htb]
\fontsize{7}{9}\selectfont
\centering
\begin{tabular}{cccccccc}
\toprule
\multicolumn{2}{c}{\multirow{2}{*}{Model}}& \multicolumn{2}{c}{Yahoo QA}& \multicolumn{2}{c}{TREC QA}&SimpleQuestions&WebQSP \\ 
\cmidrule(lr){3-4}\cmidrule(lr){5-6}\cmidrule(lr){7-7}\cmidrule(lr){8-8}
&&P@1&MRR&MAP&MRR&Accuracy&Accuracy\\
\midrule
STL&MTQA-net & 0.737 & 0.818& 0.763& 0.832&0.913&0.808\\
MTL&MTQA-net & 0.752 & 0.839 & 0.779 & 0.841&0.922 &0.820\\
\midrule   
\multirow{6}{*}{STL}&MVA-MTQA-net & 0.806 & 0.878 & 0.783 & 0.838& 0.931 & 0.823\\
&w/o word view& 0.792 & 0.863 & 0.769&0.834&0.926&0.809 \\
&w/o knowledge view& 0.781 &0.854 &0.761 &0.827 & 0.930 & 0.818 \\
&w/o semantic view& 0793&0.862&0.773&0.837&0.921&0.813 \\
&w/o knowledge semantic view& 0.788&0.859&0.762&0.822&0.928&0.814\\
&w/o co-attention view&  0.775&0.850&0.761&0.824&0.917&0.803\\
\midrule 
\multirow{6}{*}{MTL}&MVA-MTQA-net &\textbf{0.833}&	\textbf{0.909} &\textbf{0.811}&\textbf{0.862}& \textbf{0.957} & \textbf{0.858}\\
&w/o word view& 0.824 & 0.894&0.792&0.854&0.947&0.835 \\
&w/o knowledge view& 0.826 &0.893&0.796&0.861&0.944&0.844 \\
&w/o semantic view& 0.822 &0.886&0.789&0.856&0.945&0.836 \\
&w/o knowledge semantic view& 0.822 &0.890&0.793&0.856&0.944&0.840 \\
&w/o co-attention view& 0.811 &0.882&0.792&0.847&0.937&0.829 \\
\bottomrule
\end{tabular}
\caption{\label{table3} Ablation Analysis of Multi-View Attention}
\end{table*}

In this section, we conduct ablation experiments to illustrate the effect of multi-view attention scheme in the proposed method. We exclude the five kinds of views from MVA-MTQA-Net one by one and report the results in Table~\ref{table3}, including single-task learning and multi-task learning results.

From the results, we can observe that all kinds of view contribute more or less performance boost to the model. Apparently, co-attention view attention makes the most contribution to the improvement, which brings about 2-3\% increment on both STL and MTL of four tasks. 

For STL, knowledge and knowledge semantic view attentions are more distinguishable than word view and semantic view in two answer selection tasks, Yahoo QA and TREC QA, which indicates that attending more informative knowledge in the sentence is important to measure the correlation between question and answer sentence. Similarly, slight difference exists between two word-level view and two knowledge-level view attentions in SimpleQuestions and WebQSP tasks. However, the word view and semantic attentions contribute more.

For MTL, we observe that each view of attention makes a similar contribution to the improvement in four tasks, which demonstrates that the multi-view attention scheme enables each task to interact with each other in multi-task learning.

\begin{figure*}
\centering
\includegraphics[width=0.89\textwidth]{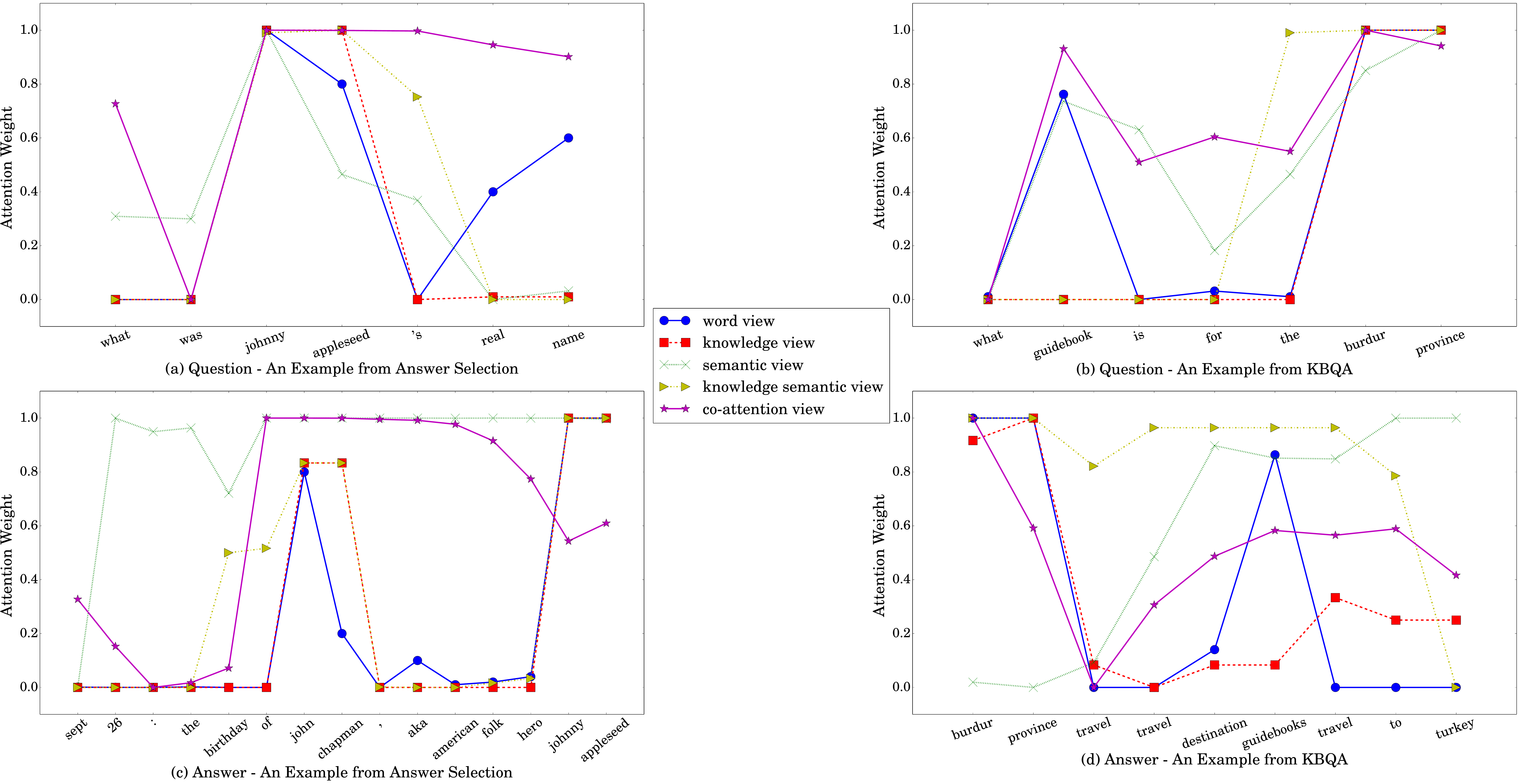}
\caption{Case Study of Multi-View Attention}
\label{figure4}
\end{figure*}

\subsection{Case Study of Multi-View Attention}
Multi-view attention scheme provides an intuitive way to inspect the importance of each word in the question and the answer by visualizing the attention weight from each kind of view. Due to the limited space, we randomly choose one question-answer pair from TREC QA dataset and one from SimpleQuestions, and visualize the attention weights predicted by MVA-MTQA-net. To be more differentiated, each view of the attention weights is normalized to [0, 1].

Figure~\ref{figure4}(a)(c) shows the predicted attention weights of an example from TREC QA. For the attention weights from the word view, we observe that words with rich information (e.g., ``johnny appleseed" and ``real name") in both question and answer receive high weights. On the other hand, entity with valuable information, such as ``johnny appleseed" and ``john chapman", are assigned with high weights in knowledge view. In two semantic views, not only similar words are attended, but also the related sentence elements, e.g., ``aka american folk hero". In co-attention view, there is a more synthetical attention distribution over all the words. In general, the distinction in word-level attentions is not as effective as that of knowledge-level. The results indicate that the incorporation of knowledge-level information aids in attending more valuable information in answer selection task.

Figure~\ref{figure4}(b)(d) provides an example from SimpleQuestions. For the question, we observe a similar distribution with the attention weights of the question example in Figure~\ref{figure4}(a), since the form of questions in KBQA task is in accord with that in answer selection task. However, for the answer, the word-level attentions compensate the insufficiency of the knowledge-level attentions, e.g., ``guidebooks", which is neglected in the knowledge-level attentions due to the incompleteness of the KG. This result demonstrates the effectiveness of combining word-level and knowledge-level information in KBQA task.

\subsection{Related Work} 
\noindent \textbf{Answer Selection} Neural networks based models has been proven effective in answer selection task, e.g., convolutional neural network (CNN)~\cite{Severyn2015Learning} and recurrent models like the long short-term memory (LSTM)~\cite{Wang2015A}.
Instead of learning the representations of the question and the answer separately, most recent studies utilize attention mechanisms to learn the interaction information between questions and answers, which can better focus on relevant parts of the input \cite{Tan2016Improved,dos2016attentive,lei2018a}. 
With the emergence of large-scale knowledge bases, several efforts have been made on incorporating external knowledge from KBs to improve answer selection models. \citeauthor{Savenkov2017EviNets} (2017) scores candidate answers by combining the supporting evidence from structured KBs and unstructured text documents. \citeauthor{Shen2018Knowledge} (2018) proposed a knowledge-aware attentive neural network to attend the knowledge information in QA sentences, based on the entities in sentences. \citeauthor{DBLP:conf/coling/DengSYLDFL18} (2018) leverage external knowledge from KB as a bridge to transfer more valuable information between cross-domain answer selection tasks.

\noindent \textbf{Knowledge Base Question Answering} Most approaches of KBQA are based on semantic parsing, which transform natural language questions into structured query to extract the answer from the knowledge base \cite{Yao2014Information,Yih2015Semantic}.
Besides, most of state-of-the-art results are achieved by deep learning models. The core idea is to learn semantic representations of both the question and the knowledge. Based on the relevancy score between the representations of the question and the knowledge, the model ultimately re-rank a list of candidate facts from KB~\cite{Bordes2015Large,Yin2016simple,Hao2017An}.
Contrary to answer selection, some recent studies integrate word or context information to aid in knowledge base question answering~\cite{yu2017improved,DBLP:conf/coling/SorokinG18}.

\noindent \textbf{Multi-Task Learning} 
Recent years have witnessed great successes of applying multi-task learning with neural networks on many NLP problems~\cite{DBLP:conf/acl/LiuQH17,guo2018soft}. Improved from the hard parameter sharing~\cite{DBLP:journals/ml/Caruana97}, most of these models consist of some task-specific private layers separated from the shared layer~\cite{DBLP:conf/icml/KumarD12}. In addition, the key innovation in most recent studies in multi-task learning is the technique used to connect the private and shared parts. \citeauthor{DBLP:conf/aaai/ChenQLH18} (2018) employs a meta-network to capture the meta-knowledge from the shared representational learning and generate the parameters of the task-specific representational learning. \citeauthor{DBLP:conf/ijcai/ZhengCQ18} (2018) exploits attention mechanisms to select the task-specific information from the shared representation layer. Different from this multi-task learning scheme, we propose a multi-view attention scheme to gather the valuable information from both task-specific and shared perspective, which enhance the interaction between different tasks.

\section{Conclusion}
In this paper, we study multi-task learning approaches to solve answer selection and knowledge base question answering simultaneously. 
We propose a novel multi-task learning scheme that utilizes multi-view attention learned from various perspectives to enable these tasks to interact with each other as well as learn more comprehensive sentence representations, including word view, knowledge view, semantic view, knowledge semantic view and co-attention view.
The experiments conducted on several widely-used benchmark QA datasets demonstrate that joint learning of answer selection and knowledge base question answering significantly outperforms single-task learning methods. Also, the multi-view attention scheme is effective in assembling attentive information from different representational perspectives to improve overall representational learning. 

\section{Acknowledgments}
This work was financially supported by the National Natural Science Foundation of China (No.61602013), the Natural Science Foundation of Guangdong (No.2018A030313017), the Shenzhen Fundamental Research Project (Grant No. JCYJ20170818091546869), and the project "PCL Future Regional Network Facilities for Large-scale Experiments and Applications". Min Yang was sponsored by CCF-Tencent Open Research Fund.

\bibliographystyle{aaai}
\bibliography{main.bib}

\end{document}